\documentclass{article}

\PassOptionsToPackage{numbers}{natbib}
\usepackage[final]{include/neurips_2019}

\usepackage[utf8]{inputenc} 
\usepackage[T1]{fontenc}    
\usepackage{hyperref}       
\usepackage{url}            
\usepackage{booktabs}       
\usepackage{amsfonts}       
\usepackage{nicefrac}       
\usepackage{xfrac}       
\usepackage{microtype}      
\usepackage{wrapfig}

\usepackage{pgfplotstable}
\usepackage{pgfplots}
\usepackage{collcell}
\usepackage{xstring}
\usepackage{tabularx}
\usepackage{graphicx}
\usepackage{tabu}
\usepackage{environ}
\usepackage{fp}
\usepackage{upgreek}
\usepackage{amsmath}
\usepackage{amsthm}
\usepackage{enumitem}
\usepackage{bm}
\usepackage{subfiles}
\usepackage{array,multirow}
\usepackage{caption,subcaption}
\usepackage{makecell}
\usepackage{tikz, color}
\usetikzlibrary{matrix}
\usetikzlibrary{calc}
\usetikzlibrary{positioning}
\usetikzlibrary{graphs}
\usetikzlibrary{shapes.geometric}

\usepackage{xcolor,colortbl}
\definecolor{matlab-blue}{rgb}{0,0.4470,0.7410}
\definecolor{matlab-orange}{rgb}{0.8500,0.3250,0.0980}
\definecolor{matlab-yellow}{rgb}{0.9290,0.6940,0.1250}

\definecolor{matlab-green}{rgb}{0.4660,0.6740,0.1880}

\definecolor{matlab-red}{rgb}{0.6350,0.0780,0.1840}

\title{A Systematic Comparison of Bayesian Deep Learning Robustness in Diabetic Retinopathy Tasks}

\author{%
  Angelos Filos, \hspace{0.25em}
  Sebastian Farquhar, \hspace{0.25em}
  Aidan N. Gomez, \hspace{0.25em}
  Tim G. J. Rudner, \\
  \AND
  Zachary Kenton, \hspace{0.25em}
  Lewis Smith, \hspace{0.25em}
  Milad Alizadeh, \hspace{0.25em}
  Arnoud de Kroon, \hspace{0.25em}
  Yarin Gal \\
  University of Oxford \\
  \texttt{\{angelos.filos, sebastian.farquhar, yarin\}@cs.ox.ac.uk}
}

\begin{document}

\maketitle
\vspace{-6mm}
\begin{abstract}
Evaluation of Bayesian deep learning (BDL) methods is challenging. We often seek to evaluate the methods' robustness and scalability, assessing whether new tools give `better' uncertainty estimates than old ones. These evaluations are paramount for practitioners when choosing BDL tools on-top of which they build their applications.
Current popular evaluations of BDL methods, such as the UCI experiments, are lacking: Methods that excel with these experiments often fail when used in application such as medical or automotive, suggesting a pertinent need for new benchmarks in the field.
We propose a new BDL benchmark with a diverse set of tasks, inspired by a real-world medical imaging application on \emph{diabetic retinopathy diagnosis}.
Visual inputs ($512\times512$ RGB images of retinas) are considered, where model uncertainty is used for medical pre-screening---i.e.\ to refer patients to an expert when model diagnosis is uncertain. Methods are then ranked according to metrics derived from expert-domain to reflect real-world use of model uncertainty in automated diagnosis. 
We develop multiple tasks that fall under this application, including out-of-distribution detection and robustness to distribution shift. 
We then perform a systematic comparison of well-tuned BDL techniques on the various tasks.
From our comparison we conclude 
that some current techniques which solve benchmarks such as UCI `overfit' their uncertainty to the dataset---when evaluated on our benchmark these underperform in comparison to simpler baselines.
The code for the benchmark, its baselines, and a simple API for evaluating new BDL tools are made available at \url{https://github.com/oatml/bdl-benchmarks}.
\end{abstract}
\section{Introduction} \label{sec:introduction}

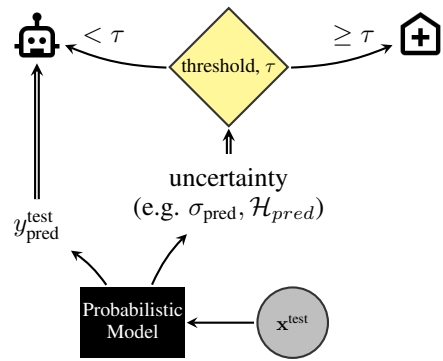
\begin{wrapfigure}{r}{0.5\textwidth}
\vspace{-21mm}
\centering
\begin{tikzpicture}[scale=0.85, >=stealth]

\tikzstyle{data}=[
  circle,
  fill =black!25,
  inner sep=1pt,
  minimum size = 12.5mm,
  thick, draw =black!80,
  node distance = 20mm,
  scale=0.75]

\tikzstyle{model}=[
  rectangle,
  fill =black,
  inner sep=1pt,
  text=white,
  minimum size = 12.5mm,
  draw=none,
  node distance = 20mm,
  scale=0.75]

\tikzstyle{thres}=[
  diamond,
  fill =yellow!50,
  inner sep=1pt,
  minimum size = 12.5mm,
  thick,
  draw =black!80,
  node distance = 20mm,
  scale=0.75]

\tikzstyle{directed}=[
  ->,
  thick,
  shorten >=0.5 pt,
  shorten <=1 pt]
  
\tikzstyle{image}=[
  inner sep=0pt]

\node[model] at (12.5,2.0) (model)     {\makecell[c]{Probabilistic\\Model}};
\node[data]  at (15.0,2.0) (x_test)    {$\mathbf{x}^{\text{test}}$};
\node        at (11.0,3.5) (y_pred)    {$y_{\text{pred}}^{\text{test}}$};
\node        at (14.0,4.0) (y_uncert)  {\makecell[c]{uncertainty\\(e.g. $\sigma_{\text{pred}}, \mathcal{H}_{pred}$)}};
\node[thres] at (14.0,6.0) (threshold) {threshold, $\tau$};
\node[image] at (11.0,6.5) (machine)   {\includegraphics[width=.05\textwidth]{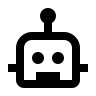}};
\node[right = 0.125cm of machine]
                           (less)      {$< \tau$};
\node[image] at (17.0,6.5) (hospital)  {\includegraphics[width=.05\textwidth]{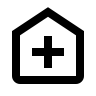}};
\node[left = 0.125cm of hospital]
                           (more)      {$\geq \tau$};

\path
        (x_test)    edge [directed]                 (model)
        (model)     edge [directed, bend right=10]  (y_pred)
        (model)     edge [directed, bend left=10]   (y_uncert)
        (y_uncert)  edge [directed, double]         (threshold)
        (threshold) edge [directed, bend left=10]   (machine)
        (threshold) edge [directed, bend right=10]  (hospital)
        (y_pred)    edge [directed, double]         (machine)
        ;
\end{tikzpicture}

\caption{Automated diagnosis: a model provides a classification and an uncertainty estimate. Predictions above an uncertainty threshold are referred to a medical expert, otherwise handled by the model.}
\label{fig:diagnosis}
\vspace{-4mm}
\end{wrapfigure}

Deep learning is continuously transforming intelligent technologies
across many fields, from advancing medical diagnostics with complex data, to
enabling autonomous driving, to deciding high-stakes economic actions
\citep{lecun2015deep}. However, deep learning models struggle to inform their
users \emph{when they don't know} -- in other words, these models fail
to communicate their uncertainty in their predictions. The implications
for deep models entrusted with life-or-death decisions are far-reaching:
experts in medical domains cannot know whether to trust their
auto-diagnostics system,
and passengers in self-driving vehicles cannot be alerted to take
control when the car does not know how to proceed.

Bayesian deep learning (BDL) offers a pragmatic approach to combining
Bayesian probability theory with modern deep learning. BDL is
concerned with the development of techniques and tools for quantifying
when deep models become uncertain, a process known as
\emph{inference} in probabilistic modelling. BDL has already been demonstrated to play
a crucial role in applications such as
medical diagnostics~\citep{leibig2017leveraging, kamnitsas2017efficient, ching2018opportunities, worrall2016automated} (see Figure~\ref{fig:diagnosis}),
computer vision~\citep{kendall2015bayesian, kendall2016modelling, kampffmeyer2016semantic},
in the sciences~\citep{levasseur2017uncertainties, mcgibbon2017improving},
and autonomous driving~\citep{amodei2016concrete, kahn2017uncertainty, kendall2015bayesian, kendall2016modelling, KendallGal2017UncertaintiesB}.

Despite BDL's impact on a range of real-world applications and
the flourish of recent ideas and inference techniques 
\citep{graves2011practical, blundell2015weight, gal2016dropout, wen2018flipout, wu2018deterministic, neklyudov2018variance},
the development of the field itself is impeded by the
lack of realistic benchmarks to guide research. Evaluating new inference
techniques on real-world applications often requires expert domain knowledge, and
current benchmarks used for the development of new inference tools
lack consideration for the cost of development, or for scalability to
real-world applications.

Advances in computer vision, natural language and reinforcement
learning are usually attributed to the emergence of challenging benchmarks,
e.g. ImageNet~\citep{deng2009imagenet}, GLUE~\citep{wang2018glue} and ALE~\citep{bellemare2013arcade}, respectively.
In contrast, many BDL papers use benchmarks such as the toy UCI datasets~\citep{hernandez2015probabilistic},
which consist of only evaluating root mean square error (RMSE) and negative log-likelihood (NLL) on
simple datasets with only a few hundred or thousand data
points, each with low input and output dimensionality. Such
evaluations are akin to toy MNIST~\citep{mnist} evaluations in deep learning.
Due to the lack of alternative standard benchmarks, in current BDL research it is common
for researchers developing new inference techniques to evaluate their methods
with such toy benchmarks alone, ignoring the demands and constraints
of the real-world applications which make use of BDL tools~\citep{mukhoti2018importance}.
This means that research in BDL broadly neglects exactly the applications that neural networks have proven themselves most effective for.

In order to make significant progress in the deployment of new
BDL inference tools, the tools must scale
to real-world settings.
And for that, researchers must be
able to evaluate their inference and iterate quickly with real-world benchmark tasks without necessarily worrying about the required
application-specific expertise.
We require benchmarks which test for inference robustness,
performance, and accuracy, in addition to cost and effort of development. 
These benchmarks should include a variety of tasks, assessing different properties of uncertainty while avoiding the pitfalls of overfitting quickly as with UCI. These should asses for scalability to large data
and be truthful to real-world applications, capturing their constraints.

\textbf{Contributions.}
We build on-top of 
previous work published at \textit{Nature Scientific Reports} by \citet{leibig2017leveraging}. We extend on their
methodology and develop an open-source benchmark, building on a downstream task which makes use of BDL in a real-world application---detecting diabetic retinopathy from fundus photos and referring the most uncertain cases for further inspection by an expert (Section~\ref{sec:benchmark}).
We extend this methodology with additional tasks that assess robustness to out-of-distribution and distribution shift, using test datasets which were collected using different medical equipment and for different patient populations.
Our implementation is easy to use for machine learning
researchers who might lack specific domain expertise, since expert details
are abstracted away and integrated into metrics which are exposed through a simple API.
Improvement on this benchmark will directly be contributing to
the advancement of an important real-world application.
We further perform a comprehensive comparison on this new benchmark, contrasting many existing BDL techniques. We develop and tune baselines for the benchmark, including Monte
Carlo dropout~\citep{gal2016dropout}, mean-field variational inference~\citep{peterson1987mean,graves2011practical, blundell2015weight}
and model ensembling~\citep{lakshminarayanan2017simple}, as well as variants of these (Section~\ref{sec:baselines}).
We conclude by demonstrating the benchmark's usefulness in ranking existing techniques
in terms of scalability and effectiveness, and show that despite the
fact that some current techniques solve benchmarks such as UCI,
they either fail to scale, fail to solve our benchmark, or fail to provide
good uncertainty estimates.
This shows that an over-reliance on UCI has the potential to badly distort work in the field because researchers prioritize their attention on approaches to Bayesian deep learning that are not suited to large scale applications (Section~\ref{sec:implications}).

It is our hope that the proposed benchmarks will make
testing new inference techniques for Bayesian deep learning radically easier, leading to faster development
iteration cycles, and rapid development of new tools.
Progress on these benchmarks will translate to more \textit{robust} and \textit{reliable} tools
for already-deployed decision-making systems, such as automatic medical diagnostics
and self-driving car prototypes.

\section{Diabetic Retinopathy Benchmark} \label{sec:benchmark}

We describe the dataset, the data processing, as well as the downstream task and metrics used.

\subsection{Dataset} \label{sub:dataset}

\begin{figure}[ht]
\centering
\begin{subfigure}[c]{1.0\linewidth}
  \includegraphics[width=0.19\linewidth]{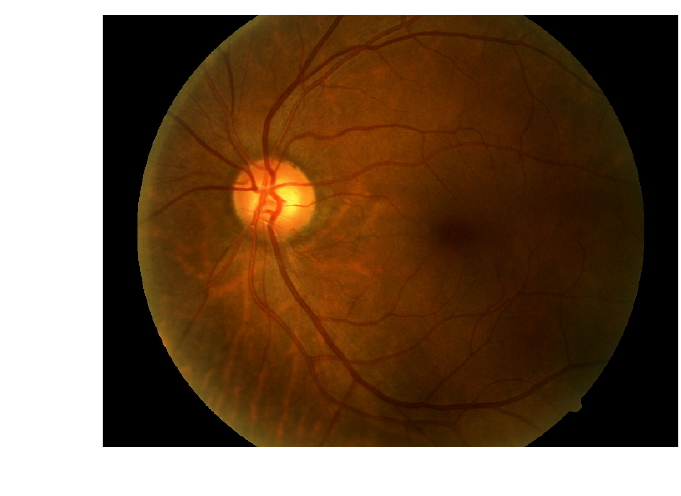}
  \includegraphics[width=0.19\linewidth]{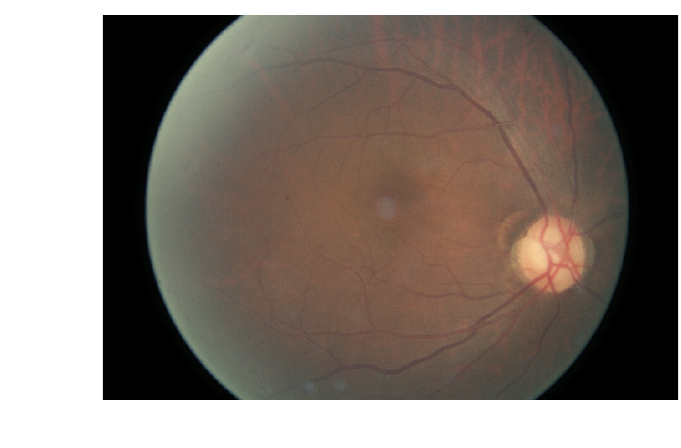}
  \includegraphics[width=0.19\linewidth]{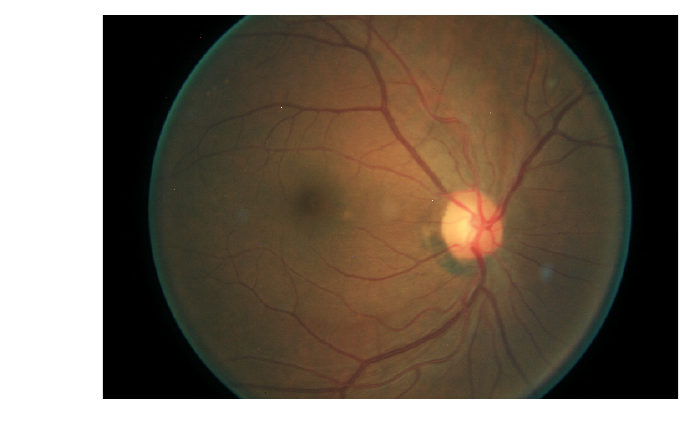}
  \includegraphics[width=0.19\linewidth]{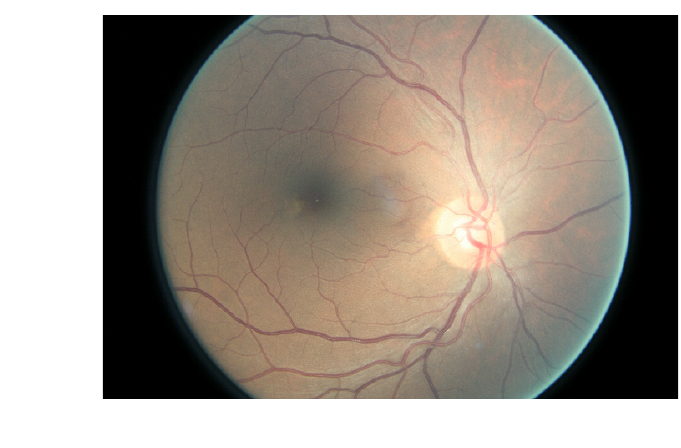}
  \includegraphics[width=0.19\linewidth]{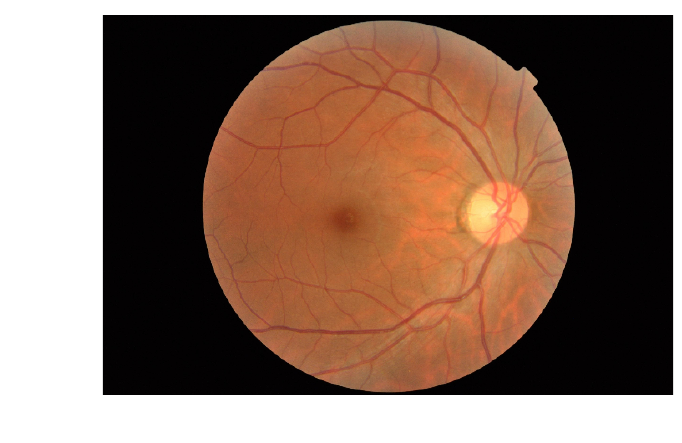}
  \caption{Healthy samples, $y=0$.}
\end{subfigure}
\begin{subfigure}[c]{1.0\linewidth}
  \includegraphics[width=0.19\linewidth]{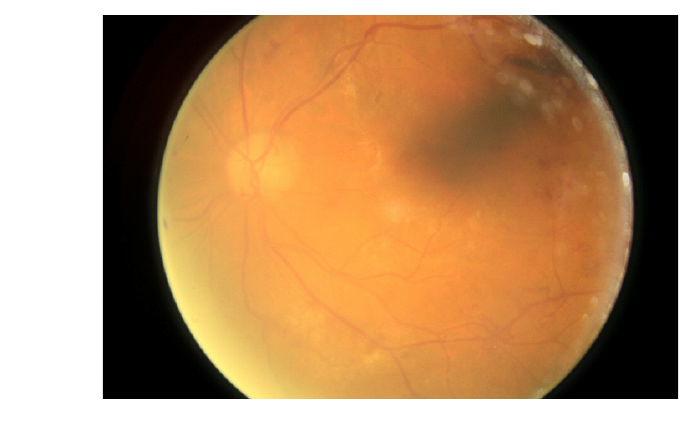}
  \includegraphics[width=0.19\linewidth]{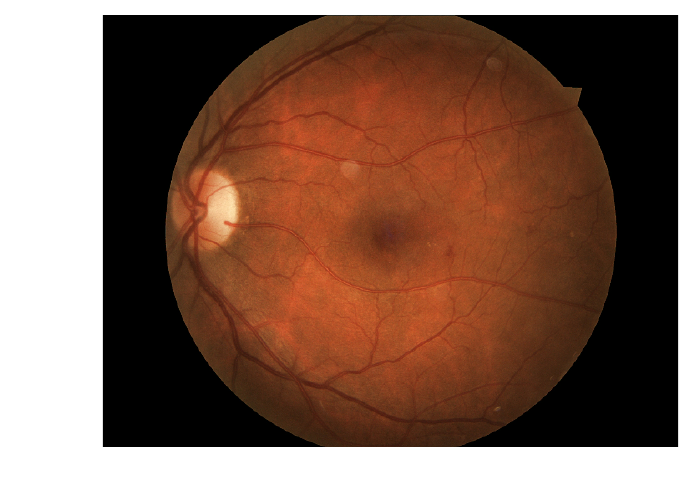}
  \includegraphics[width=0.19\linewidth]{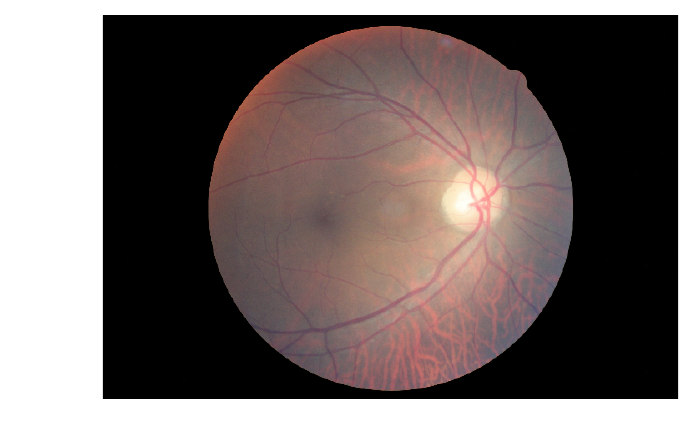}
  \includegraphics[width=0.19\linewidth]{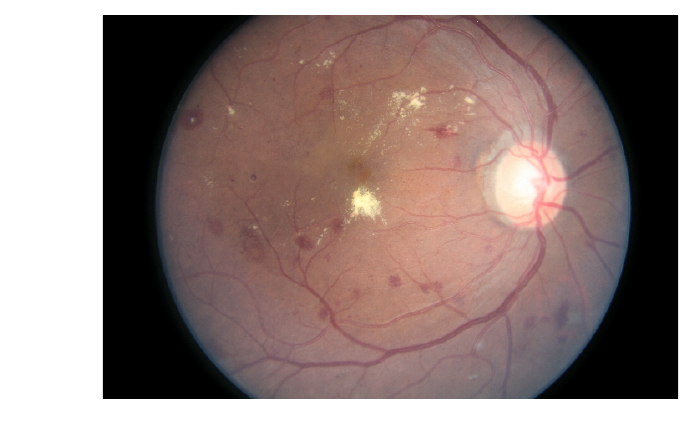}
  \includegraphics[width=0.19\linewidth]{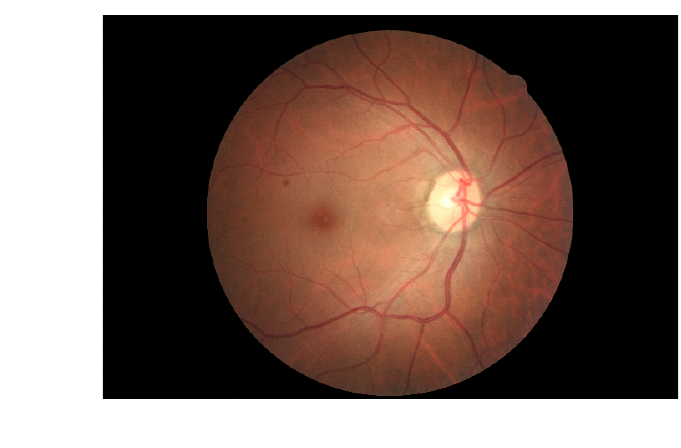}
  \caption{Unhealthy samples, $y=1$.}
\end{subfigure}
\vspace{-2mm}
\caption{Samples from the two classes, healthy and unhealthy from the raw dataset.}
\label{fig:samples}
\vspace{-4mm}
\end{figure}

The benchmark is built on the Kaggle Diabetic Retinopathy (DR)
Detection Challenge~\citep{kaggle_2015} data. It consists of 35,126 training images
and 53,576 test images. We hold-out 20$\%$ of the training data as a validation set.
Each image is graded by a specialist on the following scale: 0 -- No DR, 1 -- Mild DR, 2 -- Moderate DR, 3 -- Severe DR and
4 -- Proliferative DR. We recast the 5-class classification task as
binary classification which is easily applicable to any BDL classification algorithm by
asking the user to classify whether each image has sight-threatening
DR, which is defined as Moderate DR or greater (classes 2-4) following~\citep{leibig2017leveraging}.
Samples from both classes are provided in Figure~\ref{fig:samples}.
The data are unbalanced, with  only $19.6\%$ of the
training set and $19.2\%$ of the test set having a positive label.

Robustness to distribution shift is evaluated by training on the original
Kaggle diabetic retinopathy detection challenge dataset~\citep{kaggle_2015},
and testing on a completely disjoint APTOS 2019 Blindness Detection dataset collected in India with different medical equipment and on a different population.

\subsection{Data Processing} \label{sub:data-processing}

\begin{figure}[ht]
\centering
\begin{subfigure}[c]{1.0\linewidth}
  \includegraphics[width=0.19\linewidth]{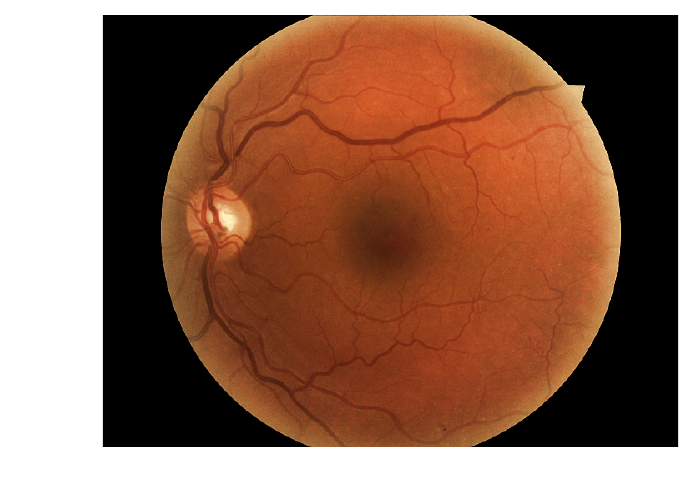}
  \includegraphics[width=0.19\linewidth]{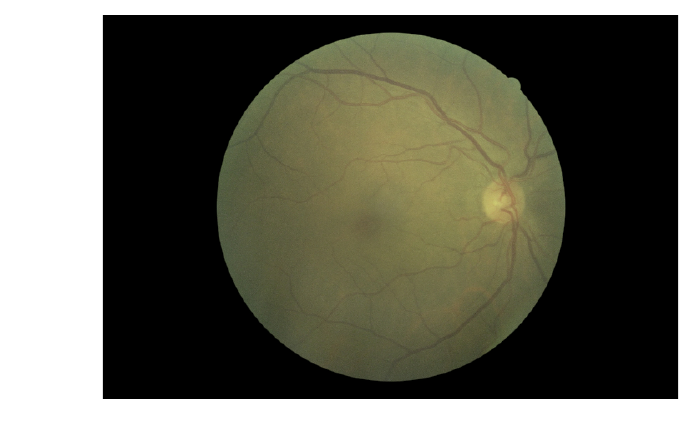}
  \includegraphics[width=0.19\linewidth]{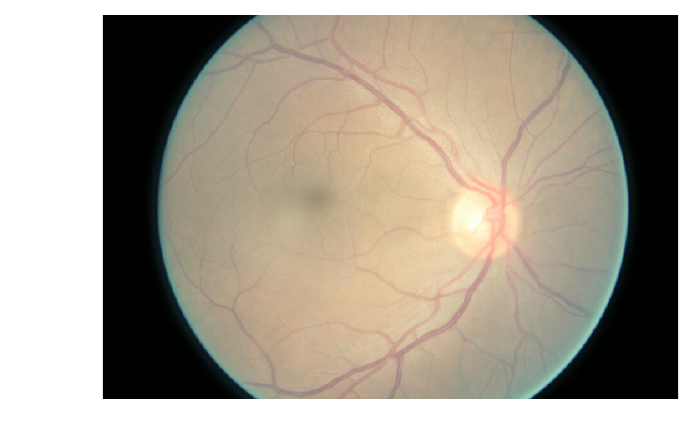}
  \includegraphics[width=0.19\linewidth]{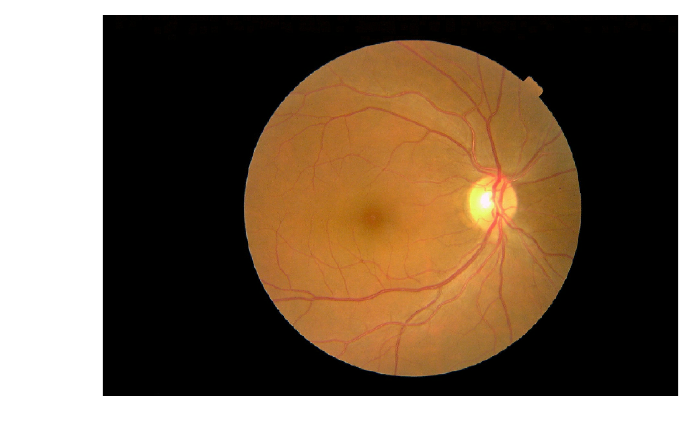}
  \includegraphics[width=0.19\linewidth]{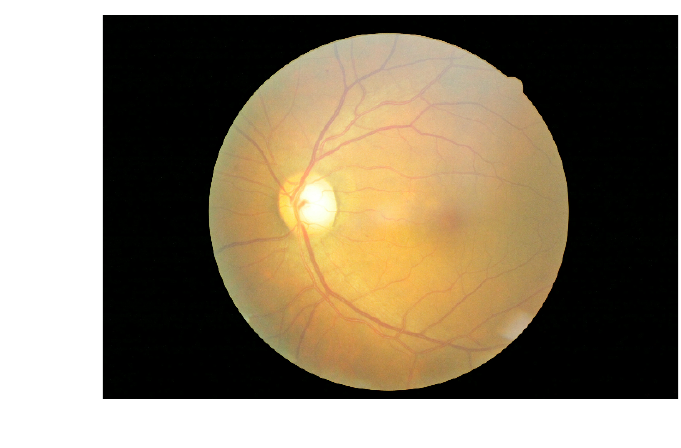}
  \caption{Original samples from the diabetic retinopathy dataset.}
\end{subfigure}
\begin{subfigure}[c]{1.0\linewidth}
  \includegraphics[width=0.19\linewidth]{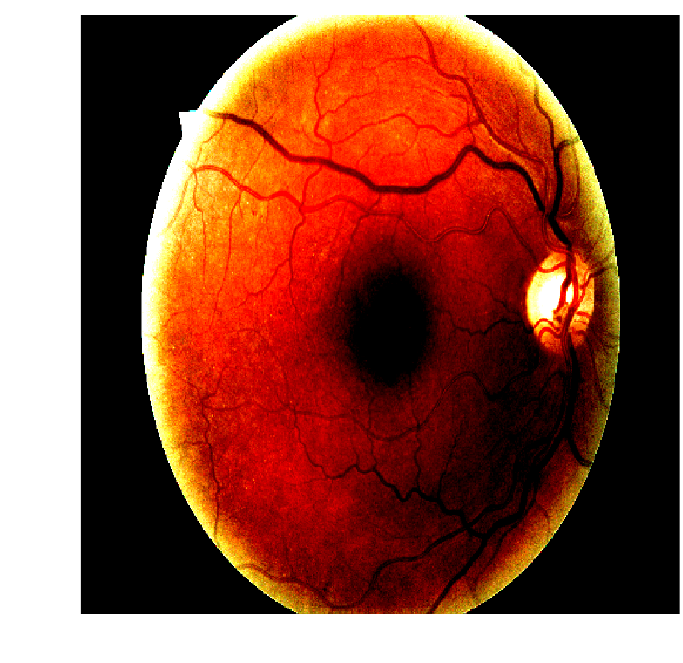}
  \includegraphics[width=0.19\linewidth]{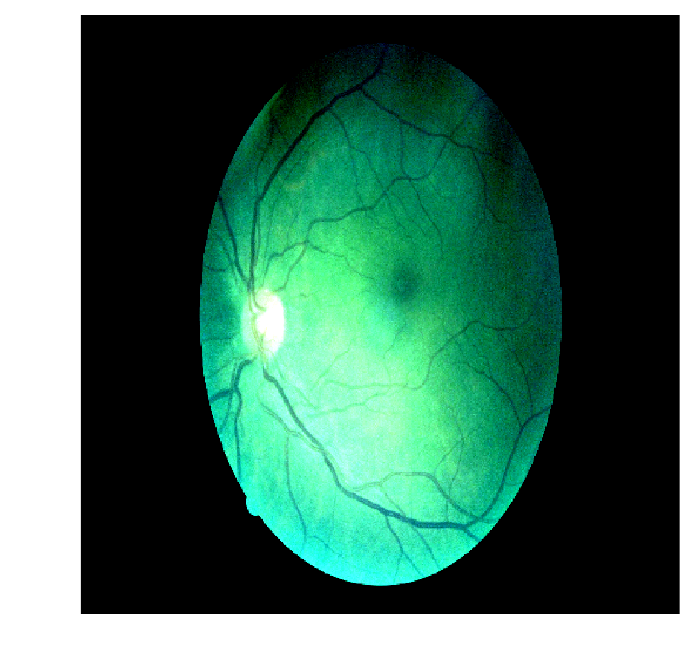}
  \includegraphics[width=0.19\linewidth]{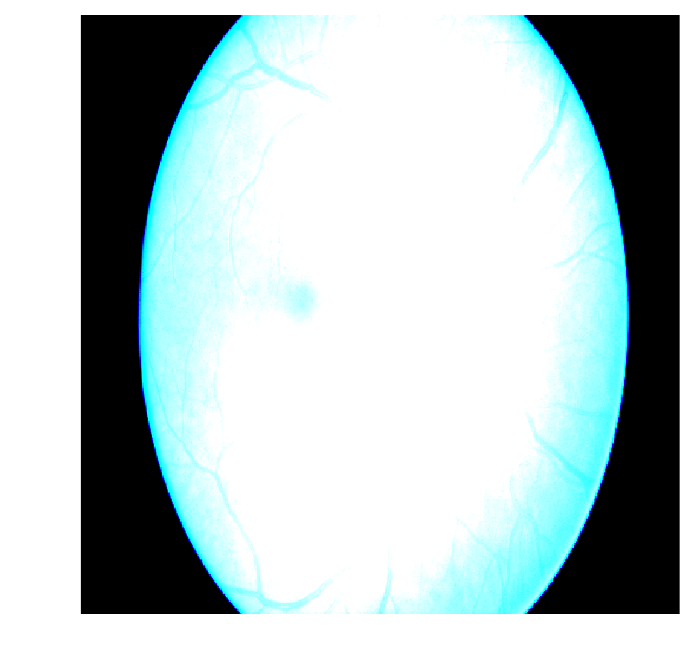}
  \includegraphics[width=0.19\linewidth]{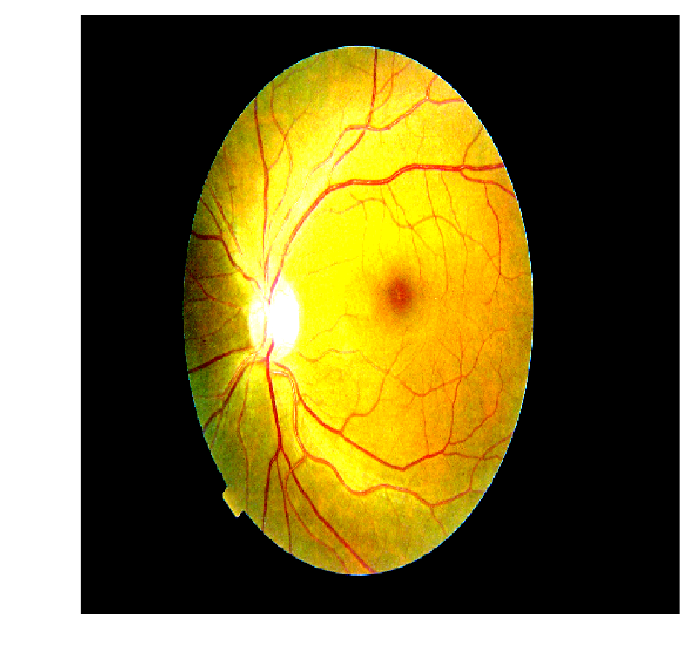}
  \includegraphics[width=0.19\linewidth]{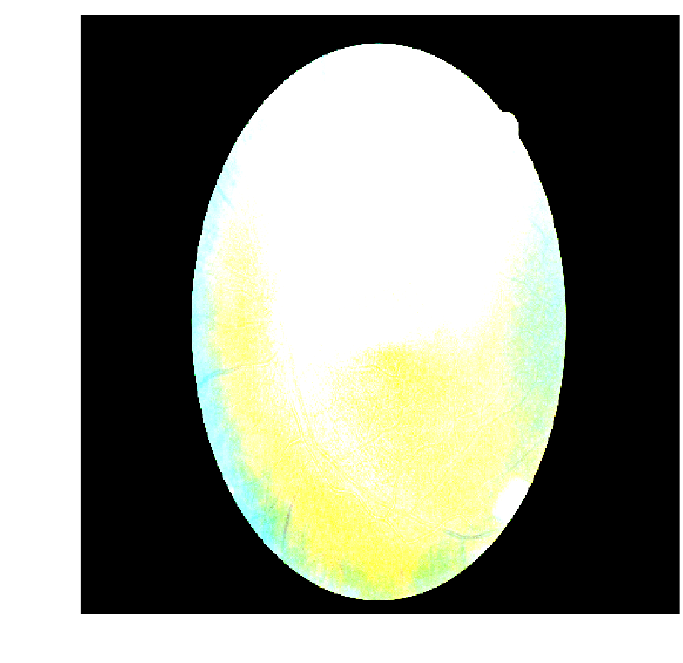}
  \caption{Processed samples from the diabetic retinopathy dataset.}
\end{subfigure}
\caption{Illustrative examples of the pre-processing procedure
applied to the original dataset.}
\label{fig:processed}
\end{figure}

All images are cropped and resized to $512\times512$, while all three colour channels are used.
The data is standard normalized for each colour channel separately, using the empirical statistics of the training data.
Similar to~\citet{leibig2017leveraging}, we augment training dataset using affine transformations,
including random zooming (by up to $\pm 10\%$), random translations (independent shifts by up to $\pm 25$ pixels)
and random rotations (by up to $\pm \pi$). Finally half of the augmented data is also flipped along the vertical and/or
the horizontal axis. Examples of original and their corresponding processed images are provided in Figure~\ref{fig:processed}.

\subsection{Downstream Task} \label{sub:downstream-task}
Machine learning researchers often evaluate their predictions directly on the whole test set.
But, in fact, in real-world settings we have additional choices available, like asking for more information when we are uncertain.
Because of the importance of accurate diagnosis, it would be unreasonable \textit{not} to ask for further scans of the most ambiguous cases.
Moreover, in this dataset, many images feature camera artefacts that distort results.
In these cases, it is critically important that a model is able to tell when the information provided to it is not sufficiently reliable to classify the patient.
Just like real medical professionals, any diagnostic algorithm should be able to flag cases that require more investigation by medical experts.
This task is illustrated in Figure~\ref{fig:diagnosis}, where a threshold, $\tau$, is used to flag cases as certain and uncertain, with uncertain cases referred to an expert. Alternatively, the uncertainty estimates could be used to come up with a priority list, which could be matched to the available resources of a hospital, rather than waste diagnostic resources on patients for whom the diagnosis is clear cut.

\begin{figure}[t]
\centering
\begin{subfigure}[l]{0.4\linewidth}
  \includegraphics[width=\linewidth]{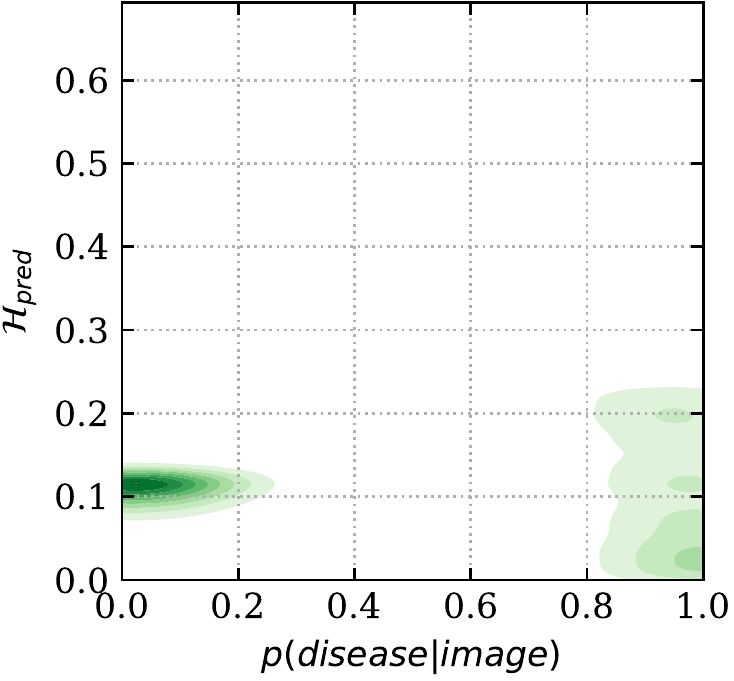}
  \caption{Correctly classified test images}
\end{subfigure}
\begin{subfigure}[l]{0.4\linewidth}
  \includegraphics[width=\linewidth]{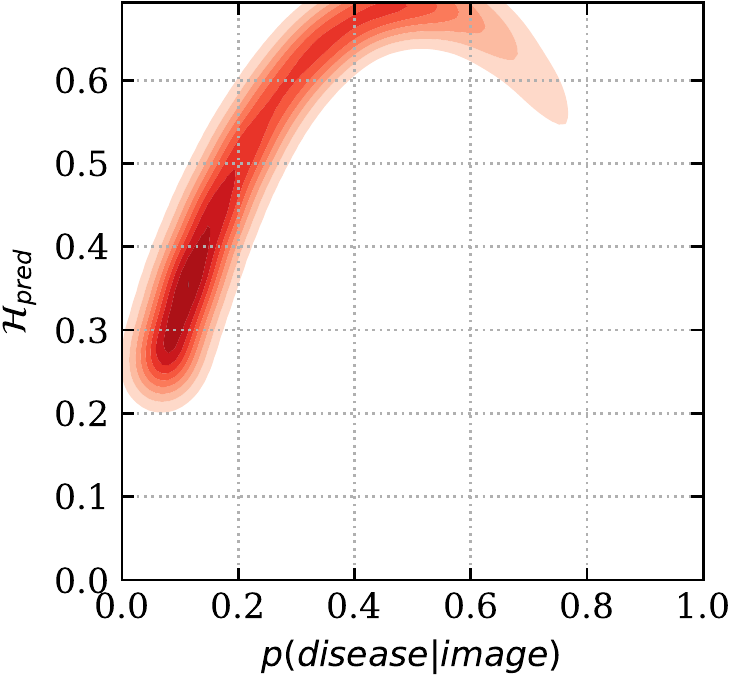}
  \caption{Erroneously classified test images}
\end{subfigure}
\caption{Relation between predictive uncertainty (i.e. entropy), $\mathcal{H}_{\text{pred}}$, of MC Dropout model,
and maximum-likelihood, i.e. sigmoid probabilities $p(\text{disease}| \text{image})$.
The model has higher uncertainty for the miss-classified images, hence
it can be used as an indicator to drive referral.}
\label{fig:kde}
\vspace{-6mm}
\end{figure}

To get some insight into the dataset, 
Figure~\ref{fig:kde} illustrates the relation between predicted probabilities, $p(\text{disease}| \text{image})$, and our estimator for
the models' uncertainty about them, the predictive entropy $\mathcal{H}_{\text{pred}}$, for an MC dropout model. Note that the model
is correct and certain about most of its predictions, as shown in sub-figure
(a), while it is more uncertain when wrong, sub-figure (b).

\subsection{Metrics} \label{sub:metrics}

In order to simulate this process of referring the uncertain
cases to experts and relying on the model's predictions for cases it is certain of, we assess the techniques by their diagnostic accuracy and
area under receiver-operating-characteristic (ROC) curve, as a function of the
referral rate. We expect the models with well-calibrated uncertainty to
refer their least confident predictions to experts (see Figure~\ref{fig:metrics}),
improving their performance as the number of referrals increases.

The accuracy of the binary classifier is defined as the
ratio of correctly classified data-points over the size of the population.
The receiver-operating-characteristic (ROC) curve (see Figure~\ref{fig:roc}) illustrates
the diagnostic ability of a binary classifier system as its discrimination threshold is varied.
It is created by plotting the true positive rate (a.k.a. sensitivity)
against the false positive rate (a.k.a. $1 - \text{sensitivity}$).
The quality of such a ROC curve can be summarized
by its area under the curve (AUC), which varies between
$0.5$ (chance level) and $1.0$ (best possible value).
\section{A Systematic Comparison of BDL Methods} \label{sec:baselines}

We next present and evaluate various Bayesian deep learning techniques
(i.e.\ baselines) on the diabetic retinopathy diagnosis benchmark. Each method
is tuned separately and, in order to obtain statistically significant
results, we train nine independent models for each method, using
different random number generator seeds. We observe consistency and robustness
for our implementations across seeds.

\paragraph{Architecture.}
Our models are deep convolutional neural networks~\citep{lecun1990handwritten}, variants of
the well-established VGG architecture~\citep{simonyan2014very} (around 2.5 million parameters).
The ADAM~\citep{kingma2014adam} adaptive optimizer with initial learning rate $4\mathrm{e}{-4}$ and
batch size $64$ are used for training all models. Leaky
rectified linear units (Leaky ReLUs)~\citep{xu2015empirical} with $\alpha = 0.2$ are
used for the hidden layers, and a sigmoid for the
output layer, modelling the probability of a patient having diabetic
retinopathy given an image of his retina, $p(\text{disease} | \text{image})$.
In contrast to~\citep{leibig2017leveraging} who uses a pre-trained network,
we initialize the weights randomly, according to~\citet{glorot2010understanding} uniform initialization method.

\paragraph{Class imbalance.}
We compensate for the class imbalance discussed in Section~\ref{sub:dataset}
by reweighing the cross-entropy part of the cost function,  attributing
more weight to the minority class, given by the relative
class frequencies in each mini-batch, $p(k)_{\text{mini-batch}}$~\citep{leibig2017leveraging}:

\begin{equation}
  \mathcal{L} = - \frac{1}{Kn} \sum_{i=1}^{n} \frac{\mathcal{L}_{\text{cross-entropy}}}{p(k)_{\text{mini-batch}}}.
\end{equation}

We also tried using a constant class weight, or artificially
balancing the two classes by sub-sampling negatively labelled images, but
both approaches made training slower and less stable for many baselines.

\paragraph{Uncertainty Estimator.}
We quantify the uncertainty of our binary classification predictions by
predictive entropy~\citep{shannon1948mathematical, gal2016uncertainty}, which captures the average amount of
information contained in the predictive distribution\footnote{The predictive uncertainty is the sum of epistemic and aleatoric uncertainty.}:

\begin{equation}
  \mathcal{H}_{\text{pred}}(y | \mathbf{x}) := - \sum_{c} p(y=c | \mathbf{x}) \log p(y=c| \mathbf{x})
\label{eq:predictive-entropy}
\end{equation}

summing over all possible classes $c$ that $y$ can take, in our case $c \in \{0, 1\}$.
This quantity is high when \textit{either} the aleatoric uncertainty is high (the input is ambiguous), \textit{or} when the epistemic uncertainty is high (a probabilistic model has many possible explanations for the input).
In practice, we approximate the $p(y=c | \mathbf{x})$ term in \eqref{eq:predictive-entropy} by $T$ Monte Carlo
samples, $\frac{1}{T} \sum_{t} p_{\theta}(y=c | \mathbf{x})$, obtained by stochastic forward passes
through the probabilistic networks. Note that this is a biased
but consistent estimator of the predictive entropy in \eqref{eq:predictive-entropy}~\citep{gal2016uncertainty}.

We contrast several methods in BDL which we discuss in more detail next.

\subsection{Bayesian Neural Networks} \label{sub:bayesian-neural-networks}

Estimating the uncertainty about a machine learning based prediction on
a single observation requires a distribution over possible outcomes, for
which a Bayesian perspective is principled. Bayesian approaches to uncertainty
estimation have indeed been proposed to assess the reliability of
clinical predictions~\citep{kononenko1993inductive, kononenko2001machine, wang2012machine, leibig2017leveraging}
but have only been applied to a handful of large-scale real-world problems~\citep{leibig2017leveraging, kahn2017uncertainty, soboczenski2018bayesian}
that neural networks (NNs) have proven themselves particularly effective for.

Finite NNs with distributions placed over the weights have been
studied extensively as Bayesian neural networks (BNNs)~\citep{neal1995bayesian, mackay1992practical, gal2016dropout},
providing robustness to over-fitting (i.e. regularization). Exact inference is analytically intractable and hence approximate inference has been applied instead
\citep{hinton1993keeping, peterson1987mean, graves2011practical, gal2016dropout}.

Given a dataset $\mathcal{D}=\{(x_{n}, y_{n})\}_{n=1}^{N}$, a BNN is defined in terms
of a prior $p(\mathbf{w})$ on the weights, as well as the likelihood $p(\mathcal{D}| \mathbf{w})$.
Variational Bayesian methods attempt to fit an approximate posterior $q(\mathbf{w})$ to
maximize the evidence lower bound (ELBO):

\begin{equation}
  \mathcal{L}_{q}=\mathbb{E}_{q}[\log p(\mathcal{D} | \mathbf{w})]-\mathrm{KL}[q(\mathbf{w}) \| p(\mathbf{w})]
\end{equation}

We parameterize $q(\mathbf{w})$ with $\theta$ parameters and choose prior distribution $p(\mathbf{w})$.
The (variational) inference is then recast as the optimization problem $\max _{\theta} \mathcal{L}_{q_{\theta}}$.
Different methods use different prior distributions and parametric families for the approximate posterior, as well as optimization methods.
We discuss these different techniques next.

\paragraph{Mean-field Variational Inference.}
Mean-field variational inference (MFVI) is an approach to learning an approximate posterior over the weights of a neural network, $q_{\theta}(\mathbf{w})$, given a prior $p(\mathbf{w})$~\citep{peterson1987mean, graves2011practical, blundell2015weight}.
In MFVI, we assume a fully-factorized Gaussian posterior (and prior).
This reduces the computational complexity of estimating the evidence lower-bound (ELBO).
In addition, we use a Monte Carlo estimate of the KL-divergence term of the ELBO in order to reduce the time complexity of a forward pass to $\mathcal{O}(D)$ in the number of weights.
\citet{blundell2015weight} applied the reparametrization trick from~\citep{kingma2013auto} to perform MFVI, which they call Bayes-by-backprop.
Instead, we use the Flipout Monte Carlo estimator of the KL-divergence~\citep{wen2018flipout}, which reduces the variance of the estimator of the gradient.
A Monte Carlo estimate of model predictions is made by taking a number of samples from the posterior distribution over the weights and averaging the predictions.

Note that the effective number of trainable parameters is doubled
compared to a deterministic NN, since both the mean and scale parameters are now learnable.
To allow fair comparison with the other baselines, we reduce the number of channels in the convolutional layers of the MFVI model to reach the model budget of 2.5 million parameters.

\paragraph{Monte Carlo Dropout.}
\citet{gal2016dropout} showed that optimising \emph{any} neural network with the
standard regularization technique of dropout~\citep{srivastava2014dropout} and
L2-regularization is equivalent to a form of variational inference in
a probabilistic interpretation of the model, so long as the dropout probability/L2 regularization are appropriately optimized~\citep{gal2016uncertainty}.
Monte Carlo samples can be drawn from the dropout NNs
by using dropout at \emph{test time}, hence the name of the method Monte Carlo Dropout (MC Dropout).
In our implementation, we chose a dropout rate to $0.2$
and perform a grid-search to set the L2-regularization coefficient.
$5\mathrm{e}{-5}$ was found to be the best value.
Better calibration of uncertainties can be obtained by optimizing the
dropout rate using convex relaxation methods as in~\citep{gal2017concrete}, but
we leave this as future work.

\subsection{Model Ensembling} \label{sub:model-ensembling}

\citet{lakshminarayanan2017simple} proposed an alternative to BNNs, termed Deep Ensembles, that
is simple to implement, readily parallelizable, requires little hyperparameter
tuning, and yields high quality predictive uncertainty estimates. The method
quantifies uncertainty by collecting predictions from $T$ independently trained deterministic
models (ensemble components). Despite the easy parallelization of the method,
the resources for training scale linearly with the required number of ensemble components $T$,
making it prohibitively expensive in some cases.

We also report results on an ensemble of  MC Dropout models, which
performs best of all the other methods, in terms of
both accuracy and AUC for all the referral rates, as illustrated in Figure~\ref{fig:metrics} and Table~\ref{tab:metrics}.
In this technique, several dropout models are separately trained in parallel.
Predictions are then made by sampling repeatedly from all models in the ensemble using fresh dropout masks for each sample, before being averaged, to form a Monte Carlo estimate of the ensemble prediction.

\subsection{Deterministic} \label{sub:deterministic}

\begin{figure}[ht]
\centering
\includegraphics[width=\linewidth]{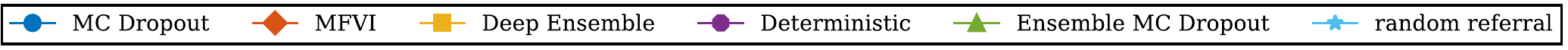}
\begin{subfigure}[l]{0.4\linewidth}
  \includegraphics[width=\linewidth]{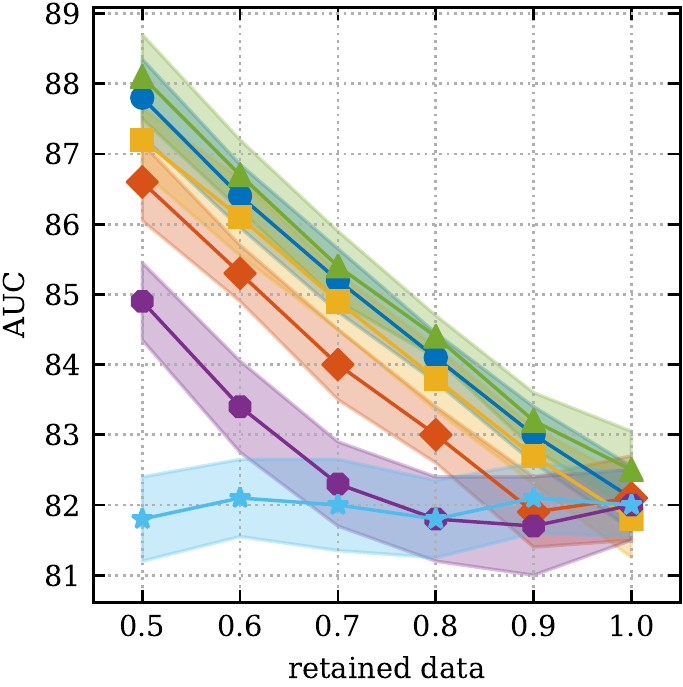}
  \caption{\textbf{test data}, AUC}
\end{subfigure}
\begin{subfigure}[r]{0.4\linewidth}
  \includegraphics[width=\linewidth]{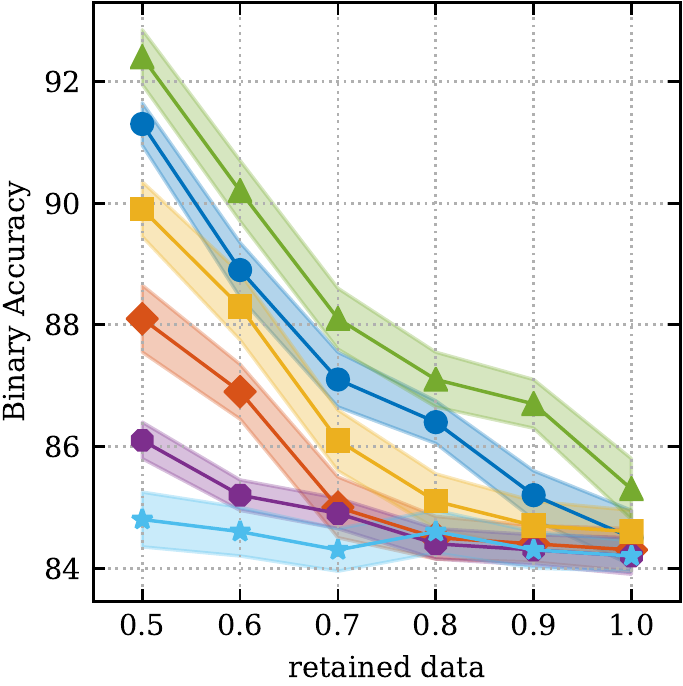}
  \caption{\textbf{test data}, accuracy}
\end{subfigure}

\vspace{1em}

\begin{subfigure}[l]{0.4\linewidth}
  \includegraphics[width=\linewidth]{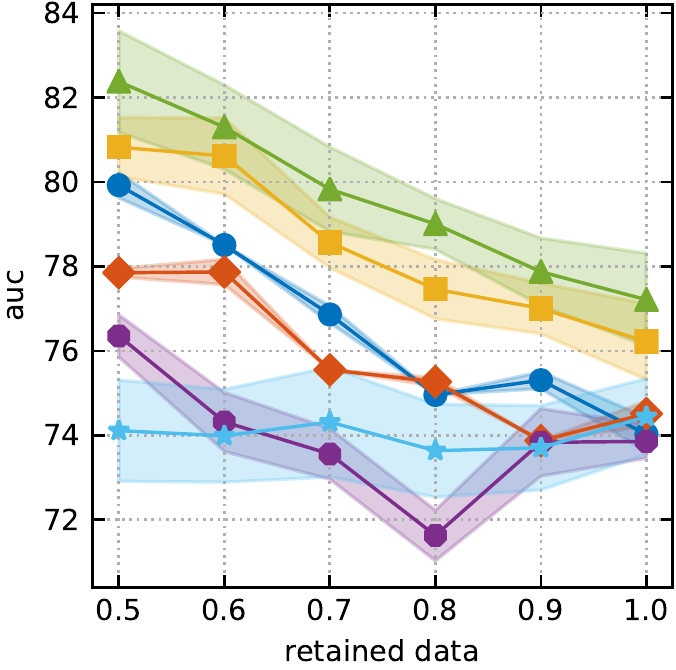}
  \caption{\textbf{distribution shift}, AUC}
\end{subfigure}
\begin{subfigure}[r]{0.4\linewidth}
  \includegraphics[width=\linewidth]{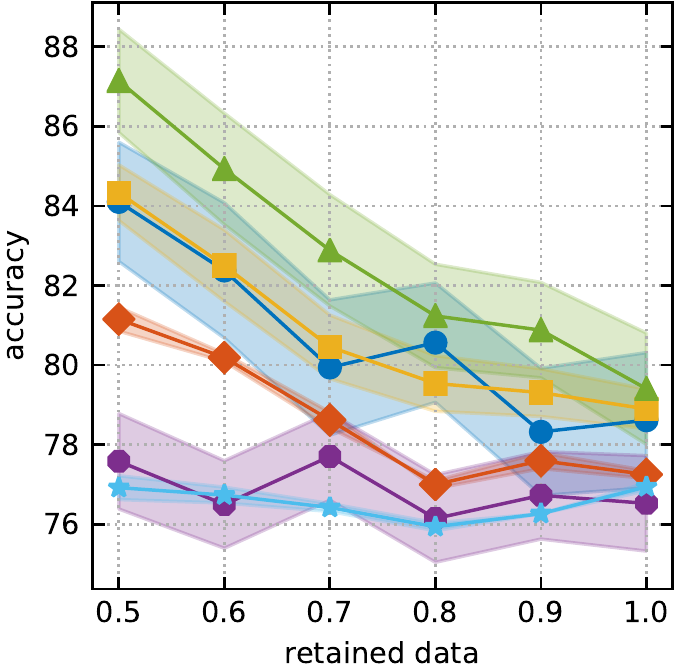}
  \caption{\textbf{distribution shift}, accuracy}
\end{subfigure}
\caption{Area under the receiver-operating characteristic curve (AUC) and binary accuracy for the
different baselines for in-distribution (a) and (b), and out-of-distribution (c) and (d) evaluation.
The methods that capture uncertainty score better when
less data is retained, referring the least certain patients to expert doctors.
The best scoring methods, \emph{MC Dropout}, \emph{mean-field variational inference} and
\emph{Deep Ensembles}, estimate and use the predictive uncertainty. The deterministic deep model regularized by \emph{standard dropout} uses only aleatoric uncertainty and performs worse.
Shading shows the standard error. The \emph{Ensemble of MC Dropout} method performs consistently better, even under the distribution shift to the APTOS 2019 dataset. However, mean-field variational inference's and MC Dropout's performance degrades in this out-of-distribution.}
\label{fig:metrics}
\vspace{-4mm}
\end{figure}

Two naive baselines are evaluated as control, a Deterministic neural network and Random.
Both are based on a deep VGG model, trained with dropout and L2-regularization, using exactly the same hyperparameters and set-up as MC Dropout.
In fact, because the conditions are identical, we used the same models for the Deterministic and MC Dropout baselines---the only difference is that for MC Dropout we sample dropout mask during evaluation and average over 100 samples from the dropout posterior to estimate uncertainty.
In contrast, the Deterministic baseline uses the sigmoid output $p(\text{disease}| \text{image})$ to quantify uncertainty, and uses the deterministic dropout approximation at test time  \cite{srivastava2014dropout}.
That is, a model is assumed to be more confident the closer to 1 or 0 its output is.
This is the simplest way a neural network might estimate uncertainty, but it captures only the aleatoric component of uncertainty---it does not capture epistemic uncertainty about the model's knowledge~\citep{gal2017what}.
Figure~\ref{fig:kde} (right)
shows that there is a correlation between the sigmoid output $p(\text{disease}| \text{image})$
and the predictive entropy $\mathcal{H}_{\text{pred}}$, which we use to measure uncertainty.
But the overall evidence in Figure~\ref{fig:metrics} and Table~\ref{tab:metrics} suggests that models which also capture the epistemic component of the uncertainty perform much better than the Deterministic baseline.

The Random baseline makes random referrals, without taking any kind of uncertainty (or input) into account.
As expected, it has the same accuracy and AUC regardless of how much data is retained vs.\ referred.

\begin{figure}[t]
\centering
\begin{subfigure}[l]{0.4\linewidth}
  \includegraphics[width=\linewidth]{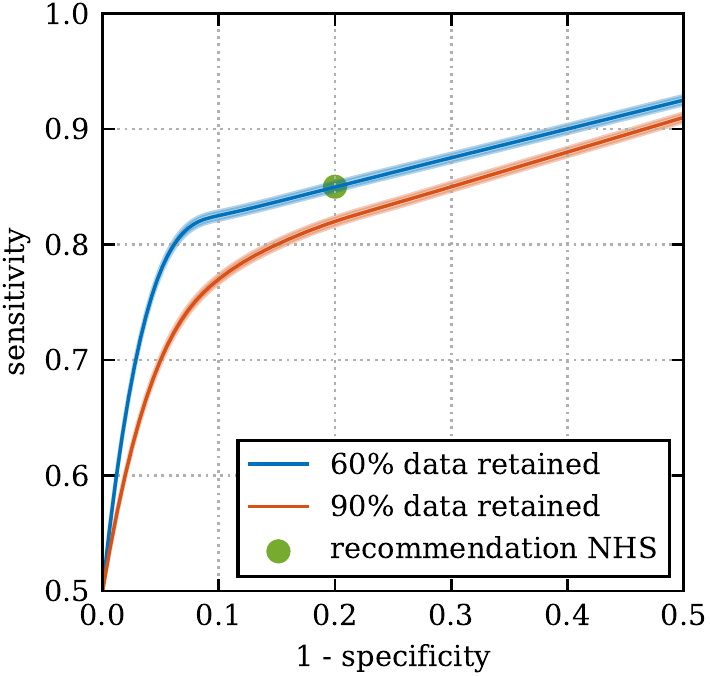}
\end{subfigure}
\begin{subfigure}[r]{0.4\linewidth}
  \includegraphics[width=\linewidth]{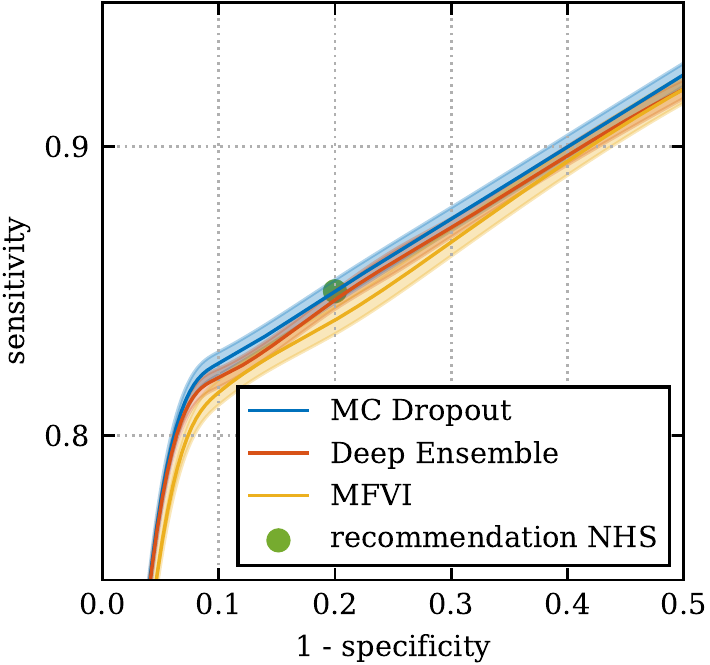}
\end{subfigure}
\caption{Receiver-operating characteristic curve (ROC) on the diabetic retinopathy diagnosis benchmark
and the NHS recommended 85\% sensitivity and 80\% specifcity ratio:
(left) the performance of MC Dropout baseline for $60\%$ and $90\%$ data retained based on predictive entropy;
(right) the comparison of the baselines at $60\%$ data retained rate. Note the different y-axes.}
\label{fig:roc}
\end{figure} 

\begin{table*}[!tb]
\small
\centering
\resizebox{\columnwidth}{!}{%
\begin{tabular}{@{\extracolsep{4pt}}lcccccc@{}}
\toprule
                        & \multicolumn{2}{c}{$50\%$ data retained} & \multicolumn{2}{c}{$70\%$ data retained} & \multicolumn{2}{c}{$100\%$ data retained} \\
                        \cline{2-3}
                        \cline{4-5}
                        \cline{6-7}\\
\textbf{Method}         &
\textbf{AUC $\uparrow$}            &
\textbf{Accuracy $\uparrow$}       &
\textbf{AUC $\uparrow$}            &
\textbf{Accuracy $\uparrow$}       &
\textbf{AUC $\uparrow$}            &
\textbf{Accuracy $\uparrow$}       \\
\midrule
\multicolumn{7}{c}{Kaggle Dataset (out-of-sample)}\\
\midrule
MC Dropout          & $\mathbf{87.8\pm1.1}$ & $\mathbf{91.3\pm0.7}$ & $\mathbf{85.2\pm0.9}$ & $87.1\pm0.9$ & $82.1\pm0.9$ & $84.5\pm0.9$ \\
Mean-field VI       & $86.6\pm1.1$ & $88.1\pm1.1$ & $84.0\pm1.0$ & $85.0\pm1.0$ & $82.1\pm1.2$ & $84.3\pm0.7$ \\
Deep Ensembles      & $87.2\pm0.9$ & $89.9\pm0.9$ & $84.9\pm0.8$ & $86.1\pm1.0$ & $81.8\pm1.1$ & $84.6\pm0.7$ \\
Deterministic       & $84.9\pm1.1$ & $86.1\pm0.6$ & $82.3\pm1.2$ & $84.9\pm0.5$ & $82.0\pm1.0$ & $84.2\pm0.6$ \\
Ensemble MC Dropout & $\mathbf{88.1\pm1.2}$ & $\mathbf{92.4\pm0.9}$ & $\mathbf{85.4\pm1.0}$ & $\mathbf{88.1\pm1.0}$ & $82.5\pm1.1$ & $85.3\pm1.0$ \\
Random              & $81.8\pm1.2$ & $84.8\pm0.9$ & $82.0\pm1.3$ & $84.3\pm0.7$ & $82.0\pm0.9$ & $84.2\pm0.5$ \\
\midrule
\multicolumn{7}{c}{APTOS 2019 Dataset (distribution shift)}\\
\midrule
MC Dropout          & $79.9\pm0.3$ & $84.1\pm1.5$ & $76.8\pm0.2$ & $79.9\pm1.7$ & $74.0\pm0.4$ & $78.6\pm1.7$ \\
Mean-field VI       & $77.8\pm0.1$ & $81.1\pm0.3$ & $75.5\pm0.0$ & $78.6\pm0.2$ & $74.5\pm0.3$ & $77.2\pm0.1$ \\
Deep Ensembles      & $80.8\pm0.7$ & $84.3\pm0.7$ & $78.5\pm0.6$ & $80.4\pm0.8$ & $76.2\pm0.9$ & $78.8\pm0.5$ \\
Deterministic       & $76.3\pm0.5$ & $77.5\pm1.2$ & $73.5\pm0.6$ & $77.7\pm1.1$ & $73.8\pm0.4$ & $76.5\pm1.2$ \\
Ensemble MC Dropout & $\mathbf{82.3}\pm1.2$ & $\mathbf{87.1}\pm1.3$ & $\mathbf{79.8\pm1.0}$ & $\mathbf{82.8\pm1.4}$ & $\mathbf{77.2\pm1.1}$ & $\mathbf{79.4\pm1.4}$ \\
Random              & $74.1\pm1.2$ & $76.9\pm1.3$ & $74.3\pm1.3$ & $76.4\pm1.0$ & $74.4\pm0.9$ & $76.9\pm0.0$ \\
\bottomrule
\end{tabular}
}
\caption{Summary performances of baselines in terms of area under the
receiver-operating-characteristic curve (AUC) and classification accuracy as a function of
retained data. In the case of no referral ($100\%$ data
retained), all methods score equally, within standard error bounds. For lower referral rates `Ensemble MC Dropout' performs best (with MC dropout matching performance in the extreme case of $50\%$ referral rate).}
\label{tab:metrics}
\vspace{-6mm}
\end{table*}

\section{Results and Analysis} \label{sec:results}

Table~\ref{tab:metrics} and Figures~\ref{fig:metrics} and~\ref{fig:roc} summarize the quantitative performance of various
methods, described in Section~\ref{sec:baselines}.
Methods that
capture meaningful uncertainty estimates show this by improving performance (i.e. AUC and
accuracy) as the rate of referral increases.
That is, steeper slopes in Figure~\ref{fig:metrics} are making better estimates of uncertainty, all else equal, because they are able to systematically refer the datapoints where their estimates are less likely to be accurate.
Note that all methods perform equally well when all data
is retained, conveying that all models have converged to
similar overall performance, providing a \emph{fair comparison of \textbf{uncertainty}}.

Benchmarks are often used to compare methodology, e.g.\ to select which tools we should build on-top. UCI, a popular benchmark in the field, has been used to reproduce such rankings of BDL methods.
Importantly, in contrast to the empirical results found in~\citep{bui2016deep} on
the toy UCI benchmark and summarised in Table~\ref{tab:uci}, our benchmark suggest a different ranking
of methods. While in~\citep{bui2016deep} mean-field variational inference outperforms the other baselines we discuss,
Table~\ref{tab:metrics} and Figures~\ref{fig:metrics} and~\ref{fig:roc} suggest that in the real-world
application of diabetic retinopathy diagnosis both ensemble methods (Section~\ref{sub:model-ensembling})
and Monte Carlo Dropout score consistently higher than MFVI, suggesting that some methods might be `overfitting' their uncertainty to the simple dataset.
That is, extensive tuning on the simple UCI tasks might have resulted in rankings which do not generalise to other tasks. Moreover, \citet{mukhoti2018importance} show that
UCI regression benchmarks are insufficient for drawing conclusions about the
effectiveness, and surely the scalability, of the inference techniques.

\begin{table*}[t]
\small
\centering
\resizebox{\columnwidth}{!}{%
\begin{tabular}{@{\extracolsep{4pt}}lccc|ccc@{}}
\toprule
                                                        &
\multicolumn{3}{c}{Log-Likelihood $\uparrow$}           &
\multicolumn{3}{c}{Root Mean Squared Error $\downarrow$}\\
                                             \cline{2-4}
                                             \cline{5-7}\\
\textbf{Datasets}                  &
\textbf{MC Dropout}                &
\textbf{Mean-field VI}             &
\textbf{Deep Ensembles}            &
\textbf{MC Dropout}                &
\textbf{Mean-field VI}             &
\textbf{Deep Ensembles}            \\
\midrule
Boston housing          & $-2.46\pm0.25$ & $-2.58\pm0.06$ & $\mathbf{-2.41\pm0.25}$
                        & $\mathbf{2.97\pm0.85}$ & $3.42\pm0.23$ & $3.28\pm1.00$ \\
Concrete                & $\mathbf{-3.04\pm0.09}$ & $-5.08\pm0.01$ & $\mathbf{-3.06\pm0.18}$
                        & $\mathbf{5.23\pm 0.53}$ & $ 5.71\pm0.15$ & $6.03\pm0.58$ \\
Energy                  & $-1.99\pm0.09$ & $\mathbf{-1.05\pm0.01}$ & $-1.38\pm0.22$
                        & $1.66\pm0.19$ & $\mathbf{0.81\pm0.08}$ & $2.09\pm0.29$ \\
Kin8nm                  & $+0.95\pm0.03$ & $+1.08\pm0.01$ & $\mathbf{+1.20\pm0.02}$
                        & $0.10\pm0.00$ & $0.37\pm0.00$ & $\mathbf{0.09\pm0.00}$ \\
Naval propulsion plant  & $+3.80\pm0.05$ & $-1.57\pm0.01$ & $\mathbf{+5.63\pm0.05}$
                        & $0.01\pm0.00$ & $0.01\pm0.00$ & $\mathbf{0.00\pm0.00}$ \\
Power plant             & $\mathbf{-2.80\pm0.05}$ & $-7.54\pm0.00$ & $\mathbf{-2.79\pm0.04}$
                        & $\mathbf{4.02\pm0.18}$ & $\mathbf{4.02\pm0.04}$ & $4.11\pm0.17 $ \\
Protein                 & $-2.89\pm0.01$ & $-3.67\pm0.00$ & $\mathbf{-2.83\pm0.02}$
                        & $\mathbf{4.36\pm0.04}$ & $4.40\pm0.02$ & $4.71\pm0.06 $ \\
Wine                    & $\mathbf{-0.93\pm0.06}$ & $-3.15\pm0.01$ & $\mathbf{-0.94\pm0.12}$
                        & $0.62\pm0.04$ & $0.65\pm0.01$ & $0.64\pm0.04$ \\
Yacht                   & $-1.55\pm0.12$ & $-4.20\pm0.05$ & $\mathbf{-1.18\pm0.21}$
                        & $\mathbf{1.11\pm0.38}$ & $1.75\pm0.42$ & $1.58\pm0.48$ \\
\bottomrule
\end{tabular}
}
\caption{Results for Deep Ensembles are borrowed from the original paper by~\citet{lakshminarayanan2017simple} and for MC Dropout and Mean-field VI from~\citep{mukhoti2018importance}.}
\label{tab:uci}
\end{table*}

\section{Implications for the Field} \label{sec:implications}
Deep learning, as a whole, has had its biggest successes when handling large, high-dimensional data.
It is something of a surprise, then, that the standard benchmarks for Bayesian deep learning, UCI, only has input dimensionalities between 4 and 16.
Due to the lack of alternative common benchmarks with well tuned baselines, researchers find it hard to publish results in Bayesian deep learning without  resorting to a comparison on UCI.
As a result, there is an undue focus in Bayesian deep learning on models that perform well with very low numbers of input features and on tiny models with a single layer of only 50 hidden units.
UCI plays an important role for a subset of models, but the fact that it is currently the field's main benchmark has a distorting effect on research.

Consider, for example, the ranking of deep learning methods for uncertainty offered by~\citet{bui2016deep}.
They compare UCI rankings from multiple papers and calculate the average rankings.
They find that Hamiltonian Monte Carlo (average rank 8.80$\pm$1.38) and mean-field variational inference (average rank 7.50$\pm$1.70) using the reparametrization trick perform best of the neural network models they consider (with the best performer being Deep Gaussian Processes).
However, HMC is known not to scale to datasets with large data, a property which is not captured with the benchmark.
Further, MC dropout is ranked second-to-last place with average rank 12.10$\pm$0.64.
Our results show that on a larger-scale dataset, MC dropout has better uncertainty estimates than mean-field variational inference and they have almost identical performance when all datapoints are retained.
Moreover, HMC would not scale to this data at all.

By relying too much on UCI as a benchmark, we give a misleading impression of relative performance, which will cause researchers to prioritise the wrong approaches.
A number of more computationally intensive extensions to MFVI have emerged since~\citet{bui2016deep} produced their analysis, while less work has gone into building on the methodology of the more computationally parsimonious Bayesian deep learning approaches like deep ensembles or MC dropout~\citep{louizos_structured_2016, sun_learning_2017, sun_functional_2019, oh_radial_2019}.
It seems likely that this is partly shaped by the fact that UCI is the predominant benchmark.

Our new benchmark and systematic comparison of BDL tools will offer a way for new methods to demonstrate their effectiveness on large-scale problems, making it easier to publish results that engage with the sorts of problems that deep learning has proven to be effective at, and which downstream users are seeking.

\bibliography{references}
\bibliographystyle{plainnat}

\end{document}